\documentclass[11pt,a4paper]{article}
\usepackage[hyperref]{emnlp-ijcnlp-2019}
\usepackage{times}
\usepackage{latexsym}

\usepackage{url}

\usepackage{graphicx}
\usepackage{multirow}
\usepackage{listings}
\usepackage{xcolor}
\usepackage{array}
\usepackage{enumitem}
\usepackage{amssymb}

\aclfinalcopy 

\title{UER: An Open-Source Toolkit for Pre-training Models}

\author{
Zhe Zhao\textsuperscript{1,2,$^{\spadesuit}$}\quad\quad
Hui Chen\textsuperscript{2,$^{\clubsuit}$}\quad\quad
Jinbin Zhang\textsuperscript{2,$^{\clubsuit}$}\quad\quad
Xin Zhao\textsuperscript{1,$^{\spadesuit}$}\quad\quad
Tao Liu\textsuperscript{1,$^{\spadesuit}$}
\\
  \textbf{
Wei Lu\textsuperscript{1,$^{\spadesuit}$}\quad\quad
Xi Chen\textsuperscript{3,$^{\blacklozenge}$}\quad\quad
Haotang Deng\textsuperscript{2,$^{\clubsuit}$}\quad\quad
Qi Ju\textsuperscript{2,$^{\clubsuit}$,\thanks{$^{*}$ Corresponding author.}{}}\quad\quad
Xiaoyong Du\textsuperscript{1,$^{\spadesuit}$}
}
\\
\textsuperscript{1} School of Information and DEKE, MOE, Renmin University of China, Beijing, China \\
\textsuperscript{2} Tencent AI Lab \\
\textsuperscript{3} School of Electronics Engineering and Computer Science, Peking University, Beijing, China \\
$^{\spadesuit}${\tt \{helloworld, zhaoxinruc, tliu, lu-wei, duyong\}@ruc.edu.cn} \\
$^{\clubsuit}${\tt \{chenhuichen, westonzhang, haotangdeng, damonju\}@tencent.com} \\
$^{\blacklozenge}${\tt \{mrcx\}@pku.edu.cn}
}

\date{}

\begin{document}
\maketitle
\begin{abstract}
  Existing works, including ELMO and BERT, have revealed the importance of pre-training for NLP tasks. While there does not exist a single pre-training model that works best in all cases, it is of necessity to develop a framework that is able to deploy various pre-training models efficiently. For this purpose, we propose an assemble-on-demand pre-training toolkit, namely Universal Encoder Representations (UER). UER is loosely coupled, and encapsulated with rich modules. By assembling modules on demand, users can either reproduce a state-of-the-art pre-training model or develop a pre-training model that remains unexplored. With UER, we have built a model zoo, which contains pre-trained models based on different corpora, encoders, and targets (objectives). With proper pre-trained models, we could achieve new state-of-the-art results on a range of downstream datasets.

\end{abstract}

\section{Introduction}

Pre-training has been well recognized as an essential step for NLP tasks since it results in remarkable improvements on a range of downstream datasets \cite{devlin2018bert}. Instead of training models on a specific task from scratch, pre-training models are firstly trained on general-domain corpora, then followed by fine-tuning on downstream tasks. Thus far, a large number of works have been proposed for finding better pre-training models. Existing pre-training models mainly differ in the following three aspects:

\hspace*{\fill}

\noindent \textbf{1) Model encoder.}

Commonly-used encoders include RNN \cite{hochreiter1997long}, CNN \cite{kim2014convolutional}, AttentionNN \cite{bahdanau2014neural}, and their combinations \cite{zhou2016attention}. Recently, Transformer (a structure based on attentionNN) is shown to be a more powerful feature extractor compared with other encoders \cite{vaswani2017attention}.

\noindent \textbf{2) Pre-training target (objective).}

Using proper target is one of the keys to the success of pre-training. While the language model is most commonly used \cite{radford2018improving}, many works focus on seeking better targets such as masked language model (cloze test) \cite{devlin2018bert} and machine translation \cite{mccann2017learned}.

\noindent \textbf{3) Fine-tuning strategy.}

Using a proper fine-tuning strategy is also important to the performance of pre-training models on downstream tasks. A commonly-used strategy is to regard pre-trained models as feature extractors \cite{kiros2015skip}.

\hspace*{\fill}

Table 1 lists 8 popular pre-training models and their main differences \cite{kiros2015skip,logeswaran2018efficient,mccann2017learned,conneau2017supervised,peters2018deep,howard2018universal,radford2018improving,devlin2018bert}. In additional to encoder, target, and fine-tuning strategy, corpus is also listed in Table 1 as an important factor for pre-training models.

\begin{table*}[!htbp]\label{models}
    \centering
\begin{small}
\newcommand{\tabincell}[2]{\begin{tabular}{@{}#1@{}}#2\end{tabular}}
        \begin{tabular}{|c|c|c|c|}

    \hline
    \textbf{Model}&\textbf{Corpus}&\textbf{Encoder}&\textbf{Target}\\
    \hline
    Skip-thoughts&Bookcorpus&GRU&Conditioned LM\\
    \hline
    Quick-thoughts&Bookcorpus+UMBCcorpus&GRU&Sentence prediction\\
    \hline
    CoVe&English-German&Bi-LSTM&Machine translation\\
    \hline
    Infersent&Natural language inference&LSTM;GRU;CNN;LSTM+Attention&Classification\\
    \hline
    ELMO&1billion benchmark&Bi-LSTM&Language model\\
    \hline
    ULMFiT&Wikipedia&LSTM&Language model\\
    \hline
    GPT&Bookcorpus; 1billion benchmark&Transformer&Language model\\
    \hline
    BERT&Wikipedia+bookcorpus&Transformer&Cloze+sentence prediction\\
    \hline
        \end{tabular}
\end{small}
\caption{\small{8 pre-training models and their differences. For space constraint of the table, fine-tuning strategies of different models are described as follows: Skip-thoughts, quick-thoughts, and infersent regard pre-trained models as feature extractors. The parameters before output layer are frozen. CoVe and ELMO transfer word embedding to downstream tasks, with other parameters in neural networks uninitialized. ULMFit, GPT, and BERT fine-tune entire networks on downstream tasks.  }}
\end{table*}

\begin{figure}[h]
\setlength{\belowcaptionskip}{-0.5cm}
\includegraphics[width=3in]{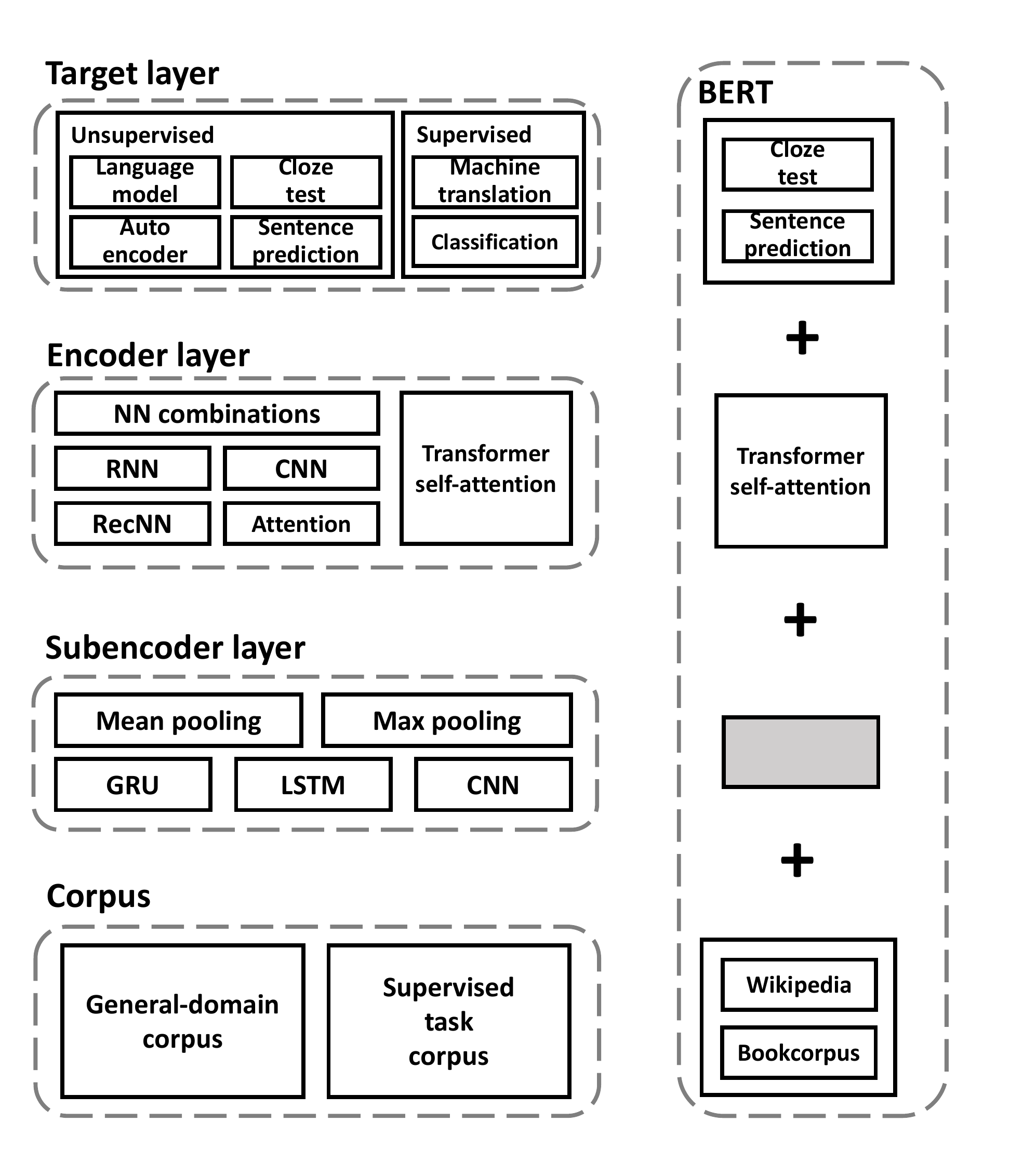}
\caption{The architecture of UER (pre-training part). We can combine modules in UER to implement BERT model.}
\end{figure}

There are many open-source implementations of pre-training models, such as Google BERT\footnote{https://github.com/google-research/bert}, ELMO from AllenAI\footnote{https://github.com/allenai/bilm-tf}, GPT and BERT from HuggingFace\footnote{https://github.com/huggingface}.
However, these works usually focus on the designs of either one or a few pre-training models.
Due to the diversity of the downstream tasks and the computational resources constraint, there does not exist a single pre-training model that works best in all cases. BERT is one of the most widely used pre-training models. It exploits two unsupervised targets for pre-training. But in some scenarios, supervised information is critical to the performance of downstream tasks \cite{conneau2017supervised,mccann2017learned}. Besides, in many cases, BERT is excluded due to its efficiency issue. Based on above reasons, it is often the case that one should adopt different pre-training models in different application scenarios.

In this work, we introduce UER, a general framework that is able to facilitate the developments of various pre-training models. UER maintains model modularity and supports research extensibility. It consists of 4 components: subencoder, encoder, target, and downstream task fine-tuning. The architecture of UER (pre-training part) is shown in Figure 1. Ample modules are implemented in each component. Users could assemble different modules to implement existing models such as BERT (right part in Figure 1), or develop a new pre-training model by implementing customized modules. Clear and robust interfaces allow users to assemble (or add) modules with as few restrictions as possible.

With the help of UER, we build a Chinese pre-trained model zoo based on different corpora, encoders, and targets. Different datasets have their own characteristics. Selecting proper models from the model zoo can largely boost the performance of downstream datasets. In this work, we use Google BERT as baseline model. We provide some use cases that are based on UER, and the results show that our models can either achieve new state-of-the-art performance, or achieve competitive results with an efficient running speed.

UER is built on PyTorch and supports distributed training mode. Clear instructions and documentations are provided to help users read and use UER codes. The UER toolkit and the model zoo are publicly available at \url{https://github.com/dbiir/UER-py}.

\section{Related Work}

\subsection{Pre-training for deep neural networks}
Using word embedding to initialize neural network's first layer is one of the most commonly used strategies for NLP tasks \cite{mikolov2013distributed,kim2014convolutional}. Inspired by the success of word embedding, some recent works try to initialize entire networks (not just first layer) with pre-trained parameters \cite{howard2018universal,radford2018improving}. They train a deep neural network upon large corpus, and fine-tune the pre-trained model on specific downstream tasks. One of the most influential works among them is BERT \cite{devlin2018bert}. BERT extracts text features with 12/24 Transformer layers, and exploits masked language model task and sentence prediction task as training targets (objectives). The drawback of BERT is that it requires expensive computational resources. Thankfully, Google makes its pre-trained models publicly available. So we can directly fine-tune on Google's models to achieve competitive results on many NLP tasks.

\subsection{NLP toolkits}

Many NLP models have tens of hyper-parameters and various tricks, and some of which exert large impacts on final performance. Sometimes it is unlikely to report all details and their effects in research paper. This may lead to a huge gap between research papers and code implementations. To solve the above problem, some works are proposed to implement a class of models in a framework. This type of work includes OpenNMT \cite{klein2017opennmt}, fairseq \cite{ott2019fairseq} for neural machine translation; glyph \cite{zhang2017encoding} for classification; NCRF++ \cite{yang2018ncrf++} for sequence labeling; Hyperwords \cite{levy2015improving}, ngram2vec \cite{zhao2017ngram2vec} for word embedding, to name a few.

Recently, we witness many influential pre-training works such as GPT, ULMFiT, and BERT. We think it could be useful to develop a framework to facilitate reproducing and refining those models. UER provides the flexibility of building pre-training models of different properties.

\section{Architecture}

In this section, we firstly introduce the core components in UER and the modules that we have implemented in each component. Figure 1 illustrates UER's framework and detailed modules (pre-training part). Modularity design of UER largely facilitates the use of pre-training models. At the end of this section, we will give some case studies to illustrate how to use UER effectively.

\subsection{Subencoder}
This layer learns word vectors from subword features. For English, we use character as subword features. For Chinese\footnote{We don't do word segmentation on Chinese corpus. We regard each Chinese character as a word. Internal structures such as radical and pinyin are regarded as Chinese subword features.}, we use radical and pinyin as subword features. As a result, the model can be aware of internal structures of words. Subword information has been explored in many NLP tasks such as text classification \cite{zhang2017encoding} and word embedding \cite{joulin2016bag}. In the pre-training literature, ELMO exploits subencoder layer. In UER, we implement RNN and CNN as subencoders, and use mean pooling or max pooling upon hidden states to obtain fixed-length word vectors.

\subsection{Encoder}

This layer learns features from word vectors. UER implements a series of basic encoders, including LSTM, GRU, CNN, GatedCNN, and AttentionNN. Users can use these basic encoders directly, or use their combinations. The output of an encoder can be fed into another encoder, forming networks of arbitrary layers. UER provides ample examples of combining basic encoders (e.g. CNN + LSTM). Users can also build their custom combinations with basic encoders in UER.

Currently, Transformer (a structure based on multi-headed self-attention) becomes a popular text feature extractor and is proven to be effective for many NLP tasks. We implement Transformer module and integrate it into UER. With Transformer module, we can implement models such as GPT and BERT easily.

\subsection{Target (objective)}
Using suitable target is the key to the success of pre-training. Many papers in this field propose
their targets and show their advantages over other ones. UER consists of a range of targets. Users
can choose one of them, or use multiple targets and give them different weights. In this section we introduce targets implemented in UER.

\begin{itemize}
\item \textbf{Language model (LM).} Language model is one of the most commonly used targets. It trains model to make it useful to predict current word given previous words.
\item \textbf{Masked LM (MLM, also known as cloze test).} The model is trained to be useful to predict masked word given surrounding words. MLM utilizes both left and right contexts to predict words. LM only considers the left context.
\item \textbf{Autoencoder (AE).} The model is trained to be useful to reconstruct input sequence as close as possible.
\end{itemize}

The above targets are related with word prediction. We call them word-level targets. Some works show that introducing sentence-level task into targets can benefit pre-training models \cite{logeswaran2018efficient,devlin2018bert}.

\begin{itemize}
\item \textbf{Next sentence prediction (NSP).} The model is trained to predict if the two sentences are continuous. Sentence prediction target is much more efficient than word-level targets. It doesn't involve sequentially decoding of words and softmax layer over entire vocabulary.
\end{itemize}

Above targets are unsupervised tasks (also known as self-supervised tasks). However, supervised tasks can provide additional knowledge that raw corpus can not provide.

\begin{itemize}
\item \textbf{Neural machine translation (NMT).} CoVe \cite{mccann2017learned} proposes to use NMT to pre-train model. The implementation of NMT target is similar with autoencoder. Both of them involve encoding source sentences and sequentially decoding words of target sentences.

\item \textbf{Classification (CLS).} Infersent \cite{conneau2017supervised} proposes to use natural language inference task (three-way classification) to pre-train model.
\end{itemize}

Most pre-training models use above targets individually. It is worth trying to use multiple targets at the same time. Some targets are complementary to each other, e.g. word-level target and sentence-level target \cite{devlin2018bert}, unsupervised target and supervised target. In experiments section, we demonstrate that proper selection of target is important. UER provides the flexibility to users in trying different targets and their combinations.

\subsection{Fine-tuning}
UER exploits similar fine-tuning strategy with ULMFiT, GPT, and BERT. Models on downstream tasks share structures and parameters with pre-training models except that they have different target layers. The entire models are fine-tuned on downstream tasks. This strategy performs robustly in practice. We also find that feature extractor strategy produces inferior results on models such as GPT and BERT.

Most pre-training works involve 2 stages, pre-training and fine-tuning. But UER supports 3 stages: 1) pre-training on general-domain corpus; 2) pre-training on downstream dataset; 3) fine-tuning on downstream dataset. Stage 2 enables models to get familiar with the distributions of downstream datasets \cite{howard2018universal,radford2018improving}. It is also called semi-supervised fine-tuning strategy in the work of \newcite{dai2015semi} since stage 2 is unsupervised and stage 3 is supervised. 

\subsection{Case Studies}
In this section, we show how UER facilitates the use of pre-training models. First of all, we demonstrate that UER can build most pre-training models easily. As shown in the following code, only a few lines are required to construct models with the interfaces in UER.

\lstset{
    numbers=left,
    numberstyle= \tiny,
    basicstyle=\scriptsize,
    keywordstyle= \color{ blue!70},
    commentstyle= \color{red!50!green!50!blue!50},
    rulesepcolor= \color{ red!20!green!20!blue!20} ,
    escapeinside=``,
    xleftmargin=2em,xrightmargin=2em, aboveskip=1em,
    framexleftmargin=2em
}

\lstset{language=python}
\begin{lstlisting}
# Implementation of BERT.
embedding = BertEmbedding(args, vocab_size)
encoder = BertEncoder(args)
target = BertTarget(args, vocab_size)

# Implementation of GPT.
embedding = BertEmbedding(args, vocab_size)
encoder = GptEncoder(args)
target = LmTarget(args, vocab_size)

# Implementation of Quick-thoughts.
embedding = Embedding(args, vocab_size)
encoder = GruEncoder(args)
target = NspTarget(args, None)

# Implementation of InferSent.
embedding = Embedding(args, vocab_size)
encoder = LstmEncoder(args)
target = ClsTarget(args, None)
\end{lstlisting}

\noindent In practice, users can assemble different subencoder, encoder, and target modules without any code work. Users can specify modules through options \emph{--subencoder}, \emph{--encoder}, and \emph{--target}. More details are available in quickstart and instructions of UER's github project. UER provides ample modules. Users can try different module combinations according to their downstream datasets. Besides trying modules implemented by UER, users can also develop their customized modules and integrate them into UER seamlessly.

\begin{table*}[!htbp]
\begin{small}
    \centering
    \begin{tabular}{|c|c|c|c|c|c|c|c|c|}

    \hline
    \textbf{Implementation}&\textbf{SST-2}&\textbf{MRPC}&\textbf{STS-B}&\textbf{QQP}&\textbf{MNLI}&\textbf{QNLI}&\textbf{RTE}&\textbf{WNLI}\\
    \hline
    HuggingFace&93.0&83.8&89.4&90.7&84.0/84.4&89.0&61.0&53.5\\
    UER&92.4&83.0&89.3&91.0&84.0/84.0&91.5&66.8&56.3\\
    \hline
    \end{tabular}

\end{small}
\caption{The performance of HuggingFace's implementation and UER's implementation on GLUE benchmark. }
\end{table*}

\begin{table*}[!htbp]
\begin{small}
    \centering
    \begin{tabular}{|c|c|c|c|c|c|}

    \hline
    \textbf{Implementation}&\textbf{XNLI}&\textbf{LCQMC}&\textbf{MSRA-NER}&\textbf{ChnSentiCorp}&\textbf{nlpcc-dbqa}\\
    \hline
    ERNIE&77.2&87.0&92.6&94.3&94.6\\
    UER&77.5&86.6&93.6&94.3&94.6\\
    \hline
    \end{tabular}

\end{small}
\caption{The performance of ERNIE's implementation and UER's implementation on ERNIE benchmark.}
\end{table*}

\section{Experiments}
To evaluate the performance of UER, experiments are conducted on a range of datasets, each of which falls into one of four categories: sentence classification, sentence pair classification, sequence labeling, and document-based QA. BERT-base uncased English model and BERT-base Chinese model are used as baseline models. In section 4.1, UER is tested on several evaluation benchmarks to demonstrate that it can produce models as intended. In section 4.2, we apply pre-trained models in our model zoo to different downstream datasets. Significant improvements are witnessed when proper encoders and targets are selected. For space constraint, we put some contents in UER's github project, including dataset and corpus details, system speed, and part of qualitative/quantitative evaluation results.

\subsection{Reproducibility}
This section uses English/Chinese benchmarks to test BERT implementation of UER. For English, we use sentence and sentence pair classification datasets in GLUE benchmark (dev set) \cite{wang2019glue}. For Chinese, we use five datasets of different types: sentiment analysis, sequence labeling, question pair matching, natural language inference, and document-based QA (provided by ERNIE\footnote{https://github.com/PaddlePaddle/ERNIE}). Table 2 and 3 compare UER's performance to other publicly available systems. We can observe that UER could match the performance of HuggingFace's and ERNIE's implementations.
Results of HuggingFace and ERNIE are reported on their github projects.
Results of UER can be reproduced by scripts in UER's github project.

\subsection{Influence of targets and encoders}

In this section, we give some examples of selecting pre-trained models given downstream datasets. 
Three Chinese sentiment analysis datasets are used for evaluation. They are Douban book review, Online shopping review, and Chnsenticorp datasets.

First of all, we use UER to pre-train on large-scale Amazon review corpus with different targets. The parameters are initialized by BERT-base Chinese model. The target of original BERT consists of MLM and NSP. However, NSP is not suitable for sentence-level reviews (we have to split reviews into multiple parts). Therefore we remove NSP target. In addition, Amazon reviews are attached with users' ratings. To this end, we can exploit CLS target for pre-training (similar with InferSent). We fine-tune these pre-trained models (with different targets) on downstream datasets. The results are shown in Table 4. BERT baseline (BERT-base Chinese) is pre-trained upon Chinese Wikipedia. We can observe that pre-training on Amazon review corpus can improve the results significantly. Using CLS target achieves the best results in most cases.

\begin{table}[!htbp]
    \centering
\begin{footnotesize}
        \begin{tabular}{|c|c|c|c|}

    \hline
    Dataset&Douban.&Shopping.&Chn.\\
    \hline
    BERT baseline&87.5&96.3&94.3\\
    \hline
    MLM&88.1&\textbf{97.0}&95.0\\
    \hline
    CLS&\textbf{88.3}&\textbf{97.0}&\textbf{95.8}\\
    \hline
        \end{tabular}
\end{footnotesize}
\caption{Performance of pre-training models with different targets.}
\end{table}

BERT requires heavy computational resources. To achieve better efficiency, we use UER to substitute 12-layers Transformer encoder with a 2-layers LSTM encoder (embedding size and hidden size are 512 and 1024). We still use the above sentiment analysis datasets for evaluation. The model is firstly trained on mixed large corpus with LM target, and then trained on large-scale Amazon review corpus with LM and CLS targets. Table 5 lists the results of different encoders. Compared with BERT baseline, LSTM encoder can achieve comparable or even better results when proper corpora and targets are selected. 

\begin{table}[!htbp]
    \centering
\begin{footnotesize}
        \begin{tabular}{|c|c|c|c|c|}

    \hline
    Dataset&Douban.&Shopping.&Chn.\\
    \hline
    BERT baseline&\textbf{87.5}&96.3&94.3\\
    \hline
    LSTM&80.3&94.0&88.3\\
    \hline
    LSTM+pre-training &86.5&\textbf{96.9}&\textbf{94.5}\\
    \hline
        \end{tabular}
\end{footnotesize}
\caption{Performance of pre-training models with different encoders.}
\end{table}

For space constraint, this section only uses sentiment analysis datasets as examples to analyze the influence of different targets and encoders. More tasks and pre-trained models are discussed in UER's github project.

\section{Conclusion}
This paper describes UER, an open-source toolkit for pre-training on general-domain corpora and fine-tuning on downstream tasks. We demonstrate that UER can largely facilitate implementations of different pre-training models. With the help of UER, we pre-train models based on different corpora, encoders, targets and make these models publicly available. By using proper pre-trained models, we can achieve significant improvements over BERT, or achieve competitive results with an efficient training speed.

\section*{Acknowledgments}

This work is supported by National Natural Science Foundation of China Grant No.U1711262 and No.61472428, 2018 Tencent Rhino-Bird Elite Training Program. \\

\bibliography{emnlp-ijcnlp-2019}
\bibliographystyle{acl_natbib}

\appendix

\end{document}